%% file: main.tex
\documentclass[11pt,a4paper]{article}
\pdfoutput=1
\usepackage{hyperref}
\input{packages}

\input{macros}

\title{\model{}: Automatic Extraction of Results\\from Machine Learning Papers}
\author{
  Marcin Kardas\textsuperscript{1}\\\And
  Piotr Czapla\textsuperscript{2}\\\And
  Pontus Stenetorp\textsuperscript{3}\\\AND
  Sebastian Ruder\textsuperscript{4}\\\And
  Sebastian Riedel\textsuperscript{1,3}\\\And
  Ross Taylor\textsuperscript{1}\\\And
  Robert Stojnic\textsuperscript{1}\\\AND\\
  \textsuperscript{1}Facebook AI Research (FAIR), London, UK\\
  \textsuperscript{2}n-waves, Wroc{\l}aw, Poland\\
  \textsuperscript{3}Department of Computer Science, University College London (UCL), UK\\
  \textsuperscript{4}DeepMind, London, UK\\
}
\date{February 2020}

\begin{document}
\setlength{\belowcaptionskip}{-12pt}
\addtolength{\textfloatsep}{-0.15in}

\maketitle

\input{abstract}
\input{intro}

\input{pipeline}

\input{dataset}
\input{experiments}
\input{ablation}

\input{conclusions}
\input{acknowledgements}

\bibliographystyle{acl_natbib}
\bibliography{bibliography}

\input{appendix}
\end{document}

%% file: packages.tex
\usepackage[nohyperref]{emnlp2020}
\usepackage{times}
\usepackage{latexsym}

\usepackage{microtype}

\aclfinalcopy

\usepackage{lmodern}
\usepackage{amssymb}
\usepackage{mathtools}
\usepackage{graphicx}
\usepackage{booktabs}
\usepackage{multirow}
\usepackage{colortbl}
\usepackage{tikz}
\usepackage{enumerate}
\usepackage{amsfonts}
\usepackage[T1]{fontenc}
\usepackage[utf8]{inputenc}
\usepackage{xcolor}
\usepackage{algorithm}
\usepackage[noend]{algorithmic}
\usepackage{xspace}
\usepackage{graphics}
\usepackage{pgfplots}
\usepackage{tikz}
\usetikzlibrary{arrows,arrows.meta,calc,decorations.pathmorphing,decorations.pathreplacing,decorations.markings,backgrounds,positioning,fit,petri,chains, matrix,scopes,shapes.geometric,quotes}
\usetikzlibrary{pgfplots.dateplot}
\usepgfplotslibrary{dateplot}
\usepackage{pgfcalendar}
\usepackage{framed}
\usepackage[capitalise,noabbrev]{cleveref}
\usepackage{varwidth}
\usepackage{soul}
\sethlcolor{green}
\usetikzlibrary{positioning,fit,matrix}
\usepackage{subcaption}
\usepackage[disable]{todonotes}

%% file: macros.tex
\newcommand{\PrOP}{\mathbb{P}}
\newcommand{\pr}[1]{\PrOP\left(#1\right)}

\newcommand{\model}{\textsc{AxCell}}

\newcommand{\unlabeled}{\textsc{ArxivPapers}}
\newcommand{\leaderboards}{\textsc{PWC Leaderboards}}
\newcommand{\linkedresults}{\textsc{LinkedResults}}
\newcommand{\finegrained}{\textsc{SegmentedTables}}

\newcommand{\prCond}[2]{\PrOP\left(#1\,|\,#2\right)}

\newcommand{\arxiv}{arXiv.org}

\newcounter{funkcyje}
\newcounter{tempcounter}

\definecolor{dark-red}{rgb}{0.4,0.15,0.15}
\definecolor{dark-blue}{rgb}{0.15,0.15,0.4}
\definecolor{medium-blue}{rgb}{0,0,0.5}

\ifaclfinal
  \newcommand{\anonurl}[1]{\url{#1}}
  \newcommand{\anonhref}[2]{\href{#1}{#2}}
\else
  \newcommand{\anonurl}[1]{\url{https://ANONYMIZED}}
  \newcommand{\anonhref}[2]{\href{https://ANONYMIZED}{ANONYMIZED}}
\fi

%% file: abstract.tex
\begin{abstract}
 Tracking progress in machine learning has become increasingly difficult with the recent explosion in the number of papers. In this paper, we present \model{}, an automatic machine learning pipeline for extracting results from papers. \model{} uses several novel components, including a table segmentation subtask, to learn relevant structural knowledge that aids extraction. When compared with existing methods, our approach significantly improves the state of the art for results extraction. We also release a structured, annotated dataset for training models for results extraction, and a dataset for evaluating the performance of models on this task. Lastly, we show the viability of our approach enables it to be used for semi-automated results extraction in production, suggesting our improvements make this task practically viable for the first time.
  Code is available on GitHub.\footnote{\anonurl{https://github.com/paperswithcode/axcell}}
 
\end{abstract}

%% file: intro.tex
\section{Introduction}
Machine learning studies how machines learn with respect to a task, a performance metric, and a dataset~\citep{mitchell2006}.  The (task, dataset, metric name, metric value) tuple can therefore be seen as representing a single result of a machine learning paper. To make progress as a field we need to make comparisons between results achieved with different methodologies. In light of the explosion in the number of machine learning publications in recent years, such comparisons have become more difficult.\footnote{In 2019, over 33,000 machine learning papers were published on the \arxiv{} open-access e-print archive, with a year-on-year growth of around $50\%$ since 2015.} This poses serious challenges to peer review, among others. For instance, across ten language modelling papers submitted to ICLR 2018, the perplexity score of the best baseline differed by more than 50 points~\cite{Ruder2018}.

One way to deal with the deluge of papers is to develop automatic approaches for extracting results from papers and aggregating them into leaderboards.
Authors typically publish their results in a tabular format in the paper, including a selection of comparisons between their approach and past papers. Automatic extraction of result tuples from tables---and optionally metadata such as model names---enables a full comparison between published methods.

Online leaderboards for comparison have become increasingly common in the research community. But these are only available for a few tasks and do not aid the comparison of models across tasks. To fill the gap, result aggregation tools such as Papers With Code\footnote{\url{https://www.paperswithcode.com/sota}} and NLP-Progress\footnote{\url{http://nlpprogress.com/}} utilise crowdsourced community contributions to populate paper leaderboards. However, human annotation of results can be laborious and error-prone, leading to omission or misreporting of paper results. This motivates the need for a machine learning approach to create a comprehensive results resource for the field.

Existing state-of-the-art approaches for results extraction are brittle and noisy, relying on text formatting hints and tables extraction from PDF files~\citep{ibm-extraction}. In contrast, we propose \model{}, a pipeline for automatic extraction of results from machine learning papers. \model{} breaks down the results extraction task into several subtasks including table type classification, table semantic segmentation and linking results to leaderboards. We employ an ULMFiT-based classifier architecture~\citep{ulmfit} to make full use of paper and table context to interpret tabular content, and extract results accordingly. 

As a whole, this paper makes three main contributions to the literature. First, we significantly improve over the state-of-the-art for results extraction with our \model{} model. On the subset of the NLP-TDMS dataset of \citet{ibm-extraction} where \LaTeX{} code is available, our approach achieves a micro F$_1$ score of 25.8 compared to the state of the art of 7.5. Secondly, we release a structured, annotated dataset for training models for results extraction, and an evaluation dataset for evaluating the performance of models on this task. Lastly, our approach is used in an in-production setting at \href{https://paperswithcode.com}{paperswithcode.com} to semi-automatically (by aiding the human review) extract results from papers and track progress in machine learning. 

\input{related}

%% file: related.tex
\section{Related Work}

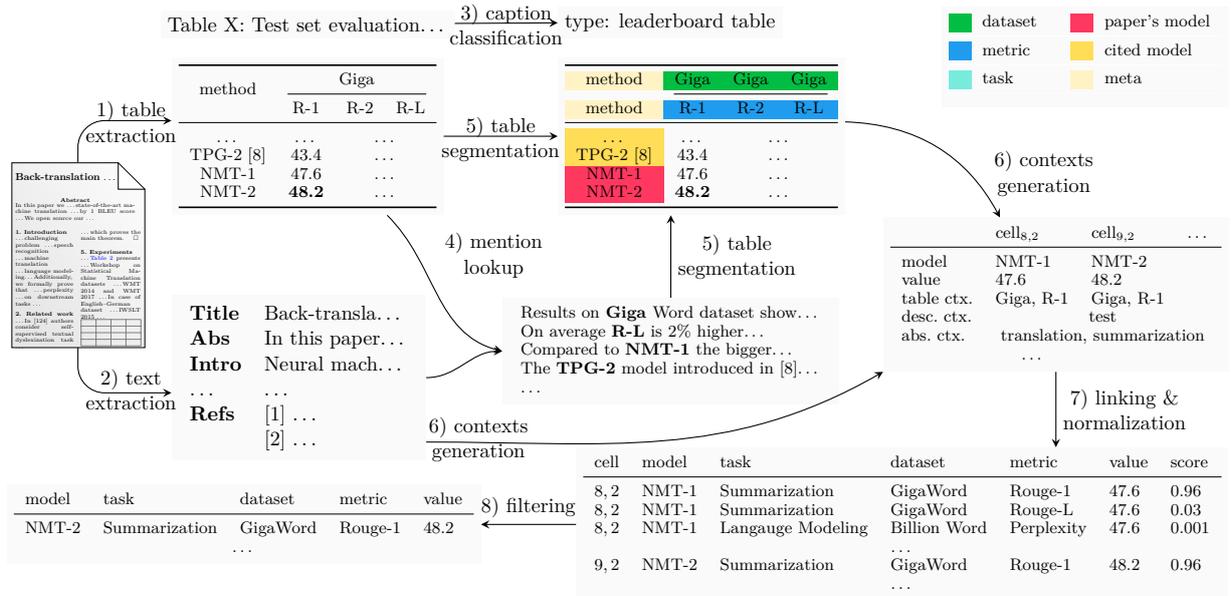
\begin{figure*}
    \centering
    \input{graphs/pipe}
    \caption{Graphical depiction of \model{}. The extraction starts with \LaTeX{} source code of a paper, from which we extract 1) tables and 2) text. 3) We classify the caption to filter out irrelevant tables. 4) The content of each cell is looked up in the paper's text. Retrieved mentions are used to 5) segment cells based on their meaning (see the legend in the top-right corner). The segmented table and the paper's text are used to 6) obtain contexts for each numeric cell. 7) Results tuples are scored based on contexts and numeric values are normalized to match required format. Finally, inferior results or results below a confidence threshold are 8) filtered out.}
    \label{fig:pipeline}
\end{figure*}

\textbf{Results Extraction}. Previous works have studied the problem of extracting results tuples (task, dataset, metric name, metric value) from papers. \citet{ranking} perform search over publications and compose a leaderboard for a queried triplet. Similar to our approach, they use tables extracted from \LaTeX{} sources. In contrast, they do not extract absolute metric values but rank papers and do not appear to utilise the text content of publications. Our goal in this paper is to extract complete results to create leaderboards, so unlike~\citet{ranking}, we focus on extracting raw metric values. Additionally we make use of the content of the publication as context for entity recognition and linking.

Closer to our formulation, \citet{ibm-extraction} extract absolute metric values alongside the metric name, task and dataset. They also use text excerpts as well as direct tabular information to make inferences for table contents.
They frame extraction as a natural language inference problem and apply an NLI model based on a BERT architecture~\citep{bert} to extract results from PDF files. The disadvantage of this approach is that using PDFs leads to a lot of noise in structural information such as the partition of a table into cells. In our work, we explicitly utilise the structural information from the \LaTeX{} source to extract entire tables in order to perform semantic segmentation. We demonstrate that this structural information and segmentation are crucial for boosting extraction performance.

\textbf{Table Extraction}. The more general problem of retrieving information from tables has been studied in past works \citep{bio-tables,tabvec,ascii-tables,tapas}. Our focus in this paper is on the problem of extracting and interpreting content of tables characteristic to machine learning papers. The goal of our table semantic segmentation model is to classify cells into categories. That is, instead of performing structural segmentation when one tries to distinguish between captions, headers and rows in a stream of text (as in~\citep{crf-tables2}) we focus on semantic segmentation (i.e., assigning roles to each cell) of tables.

%% file: graphs/pipe.tex
\def\ds#1{#1}
\def\ts#1{#1}
\def\ms#1{#1}
\definecolor{dscolor}{HTML}{02BD43}
\definecolor{tscolor}{HTML}{77ECDD}
\definecolor{mscolor}{HTML}{209CEE}
\definecolor{mtcolor}{HTML}{FFF3C5}
\definecolor{mlcolor}{HTML}{FF3860}
\definecolor{cmcolor}{HTML}{FFDD57}

\def\dscell#1{\cellcolor{dscolor}\color{black}{#1}}
\def\tscell#1{\cellcolor{tscolor}\color{white}{#1}}
\def\mscell#1{\cellcolor{mscolor}\color{black}{#1}}
\def\mtcell#1{\cellcolor{mtcolor}\color{black}{#1}}
\def\mlcell#1{\cellcolor{mlcolor}\color{black}{#1}}
\def\cmcell#1{\cellcolor{cmcolor}\color{black}{#1}}

\tikzset{>=stealth}
\tikzset{table/.style={fill=gray!4!white}}
\begin{tikzpicture}[scale=0.7, every node/.style={scale=0.7}]
    \begin{scope}[transform canvas={scale=0.5}]
    \draw[
        shading=axis,
        bottom color=black!10,
        top color=black!2,
        shading angle=-45]
        (0,0)
        --  ++(-5,0)
        --  ++(0,7)
        --  ++(4,0)
        --  ++(0,-1)
        --  ++(1,0)
        -- cycle;
    \draw (0,6) -- ++ (-1,1);
    \node[anchor=north west] (title) at (-5,6.75) {\textbf{Back-translation} \ldots};
    \node[below=0.3cm of title.south west,anchor=north west] (abs) {
        \begin{varwidth}{4.5cm}
            \begin{center}
                \tiny \textbf{Abstract}
            \end{center}
            \vspace{-0.25cm}
            \tiny In this paper we \ldots 
            state-of-the-art \ts{machine translation} \ldots
            by 1 \ms{BLEU score} \ldots We open source our \ldots
        \end{varwidth}
    };
    
    \node[below=0.05cm of abs.south west,anchor=north west] (col1) {
        \begin{varwidth}{2.25cm}
            \vspace{0.25cm}
            \tiny \textbf{1.\ Introduction}
            
            \tiny \ldots challenging problem
            \ldots \ts{speech recognition}
            \ldots \ts{machine translation} 
            \ldots \ts{language modeling}\ldots
            Additionally, we formally prove that \ldots \ms{perplexity} \ldots on downstream tasks \ldots
            
            \vspace{0.15cm}
            \tiny \textbf{2.\ Related work}
            \ldots In [124] authors consider \ts{self-supervised textual dyslexization} task \ldots 
        \end{varwidth}
    };
    
    \node[right=0.0cm of col1.north east,anchor=north west] (col2) {
        \begin{varwidth}{2.25cm}
            \tiny \ldots which proves the main theorem. \hfill\ensuremath{\square}
            \vspace{0.25cm}
            
            \tiny \textbf{5.\ Experiments}
            
            \tiny \ldots {\color{blue}Table 2} presents \ldots Workshop on Statistical \ts{Machine Translation} datasets \ldots \ds{WMT 2014} and \ds{WMT 2017} \ldots
            In case of \ds{English--German} dataset \ldots \ds{IWSLT 2015} \ldots
            
            \begin{tikzpicture}
                \draw [xstep=0.5625,ystep=0.25,gray] (0,0) grid (2.25,1);
            \end{tikzpicture}
        \end{varwidth}
    };
    \end{scope}
    
    \node[anchor=north west,table] (text) at (0.5, 1) {
        \begin{tabular}{ll}
        \textbf{Title} & Back-transla\ldots\\
        \textbf{Abs} & In this paper\ldots\\
        \textbf{Intro} & Neural mach\ldots\\
        \ldots & \ldots\\
        \textbf{Refs} & [1] \ldots\\
         & [2] \ldots\\
        \end{tabular}
    };
    
    \node[anchor=north west,table] (table) at (0.5, 5.5) {
        \small
        \begin{tabular}{cccc}
            \toprule
            \multirow{2}{*}{method} & \multicolumn{3}{c}{Giga}\\
            \cmidrule(lr){2-4}
            & R-1 & R-2 & R-L \\\midrule
            \ldots & \ldots & \multicolumn{2}{c}{\ldots}\\
            TPG-2 [8] & 43.4 & \multicolumn{2}{c}{\ldots}\\
            NMT-1 & 47.6 & \multicolumn{2}{c}{\ldots}\\
            NMT-2 & \textbf{48.2} & \multicolumn{2}{c}{\ldots}\\
            \bottomrule
        \end{tabular}
    };
    
    \node[anchor=north west,above=0.25cm of table,table] (caption) {Table X: Test set evaluation\ldots};
    
    \node[minimum width=2.5cm, minimum height=3.5cm,anchor=south east] (paper) at (0,0) {};

    \draw[->,rounded corners=10pt] (paper.south) |- ([yshift=-0.5cm]text) node[midway,above,align=center,xshift=1cm,yshift=-0.45cm] {2) text\\extraction};
    \draw[->,rounded corners=10pt] (paper.north) |- ([yshift=0.5cm]table) node[midway,below,align=center,xshift=1cm,yshift=0.5cm] {1) table\\extraction};
    
    \node[anchor=north west,table,right=1cm of text,yshift=0.5cm] (evid1) {
        \small
        \begin{tabular}{l}
        Results on \textbf{Giga} Word dataset show\ldots\\
        On average \textbf{R-L} is 2\% higher\ldots\\
        Compared to \textbf{NMT-1} the bigger\ldots\\
        The \textbf{TPG-2} model introduced in [8]\ldots\\
        \ldots\\
        \end{tabular}
    };
    
    \draw[->,in=180,out=0] (text) to (evid1.west);
    \draw[->,out=315,in=165] (table) to node[right,yshift=0.75cm,align=center] {4) mention\\lookup} (evid1.west);

    \node[anchor=north west,right=1.5cm of table,table] (segtable) {
        \small
        \begin{tabular}{cccc}
            \toprule
            \mtcell{method} & \dscell{Giga} & \dscell{Giga} & \dscell{Giga}\\
            \cmidrule(lr){2-4}
            \mtcell{method} & \mscell{R-1} & \mscell{R-2} & \mscell{R-L} \\\midrule
            \cmcell{\ldots} & \ldots & \multicolumn{2}{c}{\ldots}\\
            \cmcell{TPG-2 [8]} & 43.4 & \multicolumn{2}{c}{\ldots}\\
            \mlcell{NMT-1} & 47.6 & \multicolumn{2}{c}{\ldots}\\
            \mlcell{NMT-2} & \textbf{48.2} & \multicolumn{2}{c}{\ldots}\\
            \bottomrule
        \end{tabular}
    };
    
    \node[above=0.25cm of segtable.north west,anchor=south west,table] (tabletype) {type: leaderboard table};
    
    \draw[->] (table) -- (segtable) node[midway,below,align=center,yshift=0.5cm] {5) table\\segmentation};
    
    \draw[->] (caption) -- (tabletype) node[midway,below,align=center,yshift=0.5cm] {3) caption\\classification};
    
    \draw[->] (evid1.north) -- node[right,align=center] {5) table\\segmentation} (segtable.south -| evid1.north);
    
    \node[right=0.5cm of segtable,yshift=-3cm,table] (proposals) {
        \small
        \begin{tabular}{llll}
        & cell$_{8,2}$ & cell$_{9,2}$ & \ldots\\\midrule
          model & NMT-1 & NMT-2\\
          value & 47.6 & 48.2\\
          table ctx. & Giga, R-1 & Giga, R-1\\
          desc.\ ctx. & \multicolumn{3}{c}{test}\\
          abs.\ ctx. & \multicolumn{3}{c}{translation, summarization}\\
          \multicolumn{3}{c}{\ldots}
        \end{tabular}
    };
    
    \draw[->,bend left] (segtable) to node[right,xshift=0.95cm,yshift=-0.5cm,align=center] {6) contexts\\generation} (proposals);
    \draw[->,out=0,in=210] ([yshift=0.35cm]text.south east) to node[pos=0.1,yshift=-0.5cm,above,align=center] {6) contexts\\generation} (proposals.south west);

    \node[table,anchor=north east] (legend) at (tabletype.north -| proposals.east) {
        \small
        \begin{tabular}{lllll}
        \arrayrulecolor{white}
          \dscell{} & dataset & & \mlcell{} & paper's model\\\midrule
          \mscell{} & metric & & \cmcell{} & cited model\\\midrule
          \tscell{} & task & & \mtcell{} & meta\\\midrule
        \end{tabular}
    };

    \node[below=of proposals.south east,anchor=north east,table] (linked) {
        \small
        \begin{tabular}{lllllll}
        cell & model & task & dataset & metric & value & score\\\midrule
        ${8,2}$ & NMT-1 & Summarization & GigaWord & Rouge-1 & 47.6 & 0.96\\
        ${8,2}$ & NMT-1 & Summarization & GigaWord & Rouge-L & 47.6 & 0.03\\
        ${8,2}$ & NMT-1 & Langauge Modeling & Billion Word & Perplexity & 47.6 & 0.001 \\        \multicolumn{7}{c}{\ldots}\\
        ${9,2}$ & NMT-2 & Summarization & GigaWord & Rouge-1 & 48.2 & 0.96\\
        \multicolumn{7}{c}{\ldots}\\
        \end{tabular}
    };

    \node[left=1.25cm of linked,table] (filtered) {
        \small
        \begin{tabular}{lllll}
        model & task & dataset & metric & value\\\midrule
        NMT-2 & Summarization & GigaWord & Rouge-1 & 48.2\\
        \multicolumn{5}{c}{\ldots}\\
        \end{tabular}
    };
    
    \draw[->] (proposals.south) -- node[right,align=center] {7) linking \&\\normalization} (linked.north -| proposals.south);
    \draw[->] (linked) -- (filtered) node[midway,above] {8) filtering};
\end{tikzpicture}

%% file: pipeline.tex
\section{Our Approach}

The task of paper results extraction is to take a machine learning paper as an input and extract results contained within the paper, specifically tuples of the form (task, dataset, metric name, metric value). As an example, if we were to take in the EfficientNet paper of~\citet{efficientnet} as an input, some example results tuples we would want to extract would be EfficientNet-B7 (Image Classification, ImageNet, Top 1 Accuracy, 0.844), EfficientNet-B7 (Image Classification, ImageNet, Top 5 Accuracy, 0.971) and EfficientNet (Image Classification, Stanford Cars, Accuracy, 0.947).

To tackle this problem effectively, we need to frame the problem by defining subtasks to solve that take us from paper to results. \model{} solves several subtasks: (i) \textbf{table type classification}, identifying whether a table in a paper has relevant results; (ii) \textbf{table segmentation}, segmenting and classifying table cells according to whether they hold metrics, datasets, models, etc.; and (iii) \textbf{linking results to leaderboards}, taking the result tuples and matching them to an existing leaderboard of results. The end-to-end system is shown in~\cref{fig:pipeline} with reference to an example. We now introduce the different components of \model{}.

\subsection{Table Type Classification}
The first stage of \model{} is to categorize tables from papers into one of three categories: \texttt{leaderboard} tables, \texttt{ablation} tables and \texttt{irrelevant} tables. A \texttt{leaderboard} table contains the principal results of the paper on a selected benchmark, including comparisons with other papers. An \texttt{ablation} table compares different permutations of the paper's methodology. Lastly, \texttt{irrelevant} tables include hyperparameters, dataset statistics and other information that is not directly relevant for result extraction.

For this stage we employ a classifier with a ULMFiT architecture~\citep{ulmfit} with LSTM layers and a SentencePiece unigram model~\citep{sentencepiece} for tokenization.\footnote{Our classifier uses the fast.ai implementation~\citep{fastai}.} We train the SentencePiece model and pretrain a left-to-right ULMFiT language model on text of papers from an unlabelled dataset of arXiv articles (see \cref{sec:dataset}). \cref{tab:ulmfit} in the Appendix contains details on the hyperparameters and training regime. \footnote{We experimented with finetuning alternative language models such as BERT and SciBERT but our initial experiments did not yield superior results. A full investigation of alternative models, including pretraining from scratch, is left for future research.}

The classifier head is a standard ULMFiT classifier with a pooling layer followed by two linear layers. We treat the problem as a two-label classification with labels: \texttt{leaderboard} and \texttt{ablation}. A table is considered irrelevant if it is neither a leaderboard nor ablation (we use a confidence threshold of $0.5$). We train the model on the~\finegrained{} dataset (see~\cref{sec:dataset:structure}).

\subsection{Table Segmentation}
The second stage of \model{} is to pass relevant tables to a table segmentation subtask. The goal is to annotate each cell of a table with a label denoting what type of data a given cell contains. To this end, we classify each table cell into one of: \texttt{dataset}, \texttt{metric}, \texttt{paper model}, \texttt{cited model}, and \texttt{other} (containing meta and task cells). An example segmented table is shown in \cref{fig:pipeline}.

To help classify each table cell, we provide a context in which the cell content is mentioned. We search for cell content in the full paper content using a BM25 scoring algorithm. Retrieved text fragments are then passed to a ULMFiT-based classifier with some handcrafted features for the cell. These features include information such as the position of the cell in the table, whether the cell is a header, and cell styles. A full list is available in the Appendix. For processing the retrieved text fragments, the retrieved term from the cell is replaced with a special mask \texttt{<MASK>} token to inhibit memorization of common names (see~\cref{fig:cell-evidences} for an example). Table segmentation can then be treated as a classification problem with 5 exclusive labels. We use the same pre-trained language model weights and SentencePiece model as for the table type classification. Results for this stage of the model are outlined in \cref{tab:results:segmentation:kfolds}.

\begin{figure}
\texttt{
    On TREC-6, \textbf{<MASK>} significantly improves upon training from scratch; as examples are shorter and fewer, supervised and semi-supervised \textbf{<MASK>} achieve similar results.
}
    \caption{An example of a text excerpt from the paper by~\citet{ulmfit} used as evidence for a cell content query with \textit{ULMFiT} (covered with \texttt{<MASK>} token) as \texttt{paper model}.\\}
    \label{fig:cell-evidences}
\end{figure}

\subsection{Cell Context Generation}

The next stage after table segmentation is to generate contexts for numeric cells. As an example, if we know a numeric cell has a dataset cell somewhere in its row, and a model cell somewhere in its column, then this table context is informative for deciding the dataset and model for this result. But there is much broader context in the paper that is useful for linking.

For example, a paper studying semantic segmentation with models evaluated on KITTI and CamVid datasets could mention \textit{semantic segmentation} in the introduction, \textit{test set} in a subsection referring to a results table, \textit{KITTI} in the description of that table and \textit{class IoU} in the column header. \cref{fig:contexts} shows a visual representation of this hierarchy of context. 

To reflect this hierarchy we generate several types of contexts for each cell. The \texttt{table context}, as discussed, looks at a numeric cell and other cells in its row or column labeled as model, dataset or metric. We also define text contexts: a \texttt{caption context}, the table caption; a \texttt{mentions context}, text fragments referencing the table; an \texttt{abstract context}, the paper abstract; and a global \texttt{paper context}, containing the entire paper text. The gathered contexts are then used to link potential results to predefined leaderboards of results.

\subsection{Linking Cells to Leaderboards}
\begin{figure}
    \centering
    \input{graphs/contexts}
    \caption{Using Context Hierarchy and Evidences for Linking. This figure highlights the context hierarchy, from the global paper to the specific table, the evidence for tasks (blue), datasets (pink) and metrics (violet) for the $56.3$ value extracted from cell contexts, and lastly the result from linking.\\}
    \label{fig:contexts}
\end{figure}

Once we have the cell contexts, the next stage of \model{} is to link them to leaderboards to form performance records. The goal is to take a metric value for a model and infer the leaderboard it is connected to. A leaderboard is defined via the triplet (task, dataset, metric name). For example: (Image Classification, ImageNet, Top 1 Accuracy) can capture papers that report performance on Image Classification for ImageNet and report Top 1 Accuracy. To simplify the problem, we assume a closed-domain with all leaderboards known in advance.
To match results to leaderboards we look for evidence in cell contexts, which we now explain.

\subsubsection{Inference From Evidences}
\label{sec:linking-inference}

Pieces of evidence are words or phrases that correspond to a task, dataset or metric. For example, \textit{SST-2}, \textit{binary} and \textit{polarity} could all serve as evidence for the two-class \textit{Stanford Sentiment Treebank} dataset~\citep{sst-2}. Pieces of evidence allow us to infer whether an entity has been mentioned in a given context. Using the same example, if ``SST-2'' appears in the table caption then this is evidence that a numeric value in the table could be linked to the \textit{Stanford Sentiment Treebank} dataset.

For a given numeric cell in the table, we search the cell contexts for evidence for every entity (task, dataset, metric) and accumulate them into a set of $M$ pieces of cell evidence $E = \{e_{1}, \ldots, e_{M}\}$, with $e_{j}$ of the form $(\textit{mention}_j, \textit{entity}_j, \textit{context}_j)$. For example, (acc, metric, table) means ``acc'' metric evidence was found in cell's table context. Using this evidence set, our goal is to calculate $\prCond{y_{k}}{E}$,  where $y_{k}$ is a binary variable denoting whether the cell contains results for a leaderboard $k \in \{1, \dots, N\}$.

Through Bayes' Rule we know that $\prCond{y_{k}}{E} \propto \prCond{E}{y_{k}}\pr{y_{k}}$. We can estimate $\prCond{E}{y_{k}}$ by Naive Bayes:

$$ \prCond{E}{y_{k}} \approx \prod_{j}^{M}\prCond{e_{j}}{y_{k}} $$

Since $e_j=(\textit{mention}_j, \textit{entity}_j, \textit{context}_j)$, to model $\prCond{e_{j}}{y_{k}}$ we need to define a joint probability model for these different elements. In our results linking model, we assume a mention can appear in a given context on purpose, to describe content of the cell, or it can be noise -- falsely matched or referencing another cell. With additional simplifications we assume:

\begin{align*}
  &\prCond{e_j}{y_k} = \prCond{(\textit{mention}_j, \textit{entity}_j, \textit{ctx}_j)}{y_k}\\ &\;\propto\prCond{\textit{noise}}{\textit{ctx}_j}\cdot\prCond{\textit{entity}_j}{\textit{noise}}\\
  &\;+\prCond{\neg\textit{noise}}{\textit{ctx}_j}\cdot\prCond{\textit{mention}_j, \textit{entity}_j}{y_k}
\end{align*}
where the noise probability for each context, $\prCond{\textit{noise}}{\textit{ctx}_j}$, is computed using training set. \\

Finally, for a leaderboard $y_k=(y_k^{(\textit{task})}, y_k^{(\textit{dataset})}, y_k^{(\textit{metric})})$ we assume that $\prCond{\textit{mention}, \textit{entity}}{y_k} \!\propto\! \prCond{\textit{mention}}{y_k^{(\textit{entity})}}$, that is, a metric mention ``acc'' has the same conditional probability for leaderboard (Image Classification, ImageNet, Accuracy) as for (Natural Language Inference, SNLI, Accuracy).

We compute $\prCond{\textit{mention}}{y_k^{(\textit{entity})}}$ to be inversly proportional to the number of other entities of type \textit{entity} with the same \textit{mention} evidence (see~\cref{sec:mention-probs} for details).

\subsection{Filtering}
The final step of \model{} is to filter out results with a linking score that is too low, results for cited models and inferior results (to keep only the best performing results). First, we filter out records not associated with models introduced in a paper being processed. We then remove records for which a linking score is below some given threshold. The remaining records are grouped by leaderboard and for each leaderboard only the best result is kept, based on \textit{higher is better} annotation available in taxonomy; e.g. \textit{Accuracy} would keep higher values, \textit{Error Rate} would keep lower values. Finally, we remove all results with a linking score below the second threshold. This gives us the final list of results tuples extracted from the paper.

%% file: graphs/contexts.tex
    \def\ds#1{\colorlet{foo}{pink}\sethlcolor{foo}\hl{#1}}
    \def\ts#1{\colorlet{foo}{cyan!50}\sethlcolor{foo}\hl{#1}}
    \def\ms#1{\colorlet{foo}{magenta!80}\sethlcolor{foo}\hl{#1}}
	\begin{tikzpicture}[scale=0.7, every node/.style={scale=0.7}]
        \draw[
            shading=axis,
            bottom color=black!10,
            top color=black!2,
            shading angle=-45]
            (0,0)
            --  ++(-5,0)
            --  ++(0,7)
            --  ++(4,0)
            --  ++(0,-1)
            --  ++(1,0)
            -- cycle;
        \draw (0,6) -- ++ (-1,1);
        \node[anchor=north west] (title) at (-5,6.75) {\textbf{Back-translation} \ldots};
        \node[below=0.3cm of title.south west,anchor=north west] (abs) {
            \begin{varwidth}{4.5cm}
                \begin{center}
                    \tiny \textbf{Abstract}
                \end{center}
                
                \tiny In this paper we \ldots 
                state-of-the-art \ts{machine translation} \ldots
                by 1 \ms{BLEU score} \ldots We open source our \ldots
            \end{varwidth}
        };
        
        \node[below=0.15cm of abs.south west,anchor=north west] (col1) {
            \begin{varwidth}{2.25cm}
                \vspace{0.25cm}
                \tiny \textbf{1.\ Introduction}
                
                \tiny \ldots challenging problem
                \ldots \ts{speech recognition}
                \ldots \ts{machine translation} 
                \ldots \ts{language modeling}\ldots
                Additionally, we formally prove that \ldots \ms{perplexity} \ldots on downstream tasks \ldots
                
                \vspace{0.15cm}
                \tiny \textbf{2.\ Related work}
                \ldots In [124] authors consider \ts{self-supervised textual dyslexization} task \ldots 
            \end{varwidth}
        };
        
        \node[right=0.0cm of col1.north east,anchor=north west] (col2) {
            \begin{varwidth}{2.25cm}
                \tiny \ldots which proves the main theorem. \hfill\ensuremath{\square}
                \vspace{0.25cm}
                
                \tiny \textbf{5.\ Experiments}
                
                \tiny \ldots {\color{blue}Table 2} presents \ldots Workshop on Statistical \ts{Machine Translation} datasets \ldots \ds{WMT 2014} and \ds{WMT 2017} \ldots
                In case of \ds{English--German} dataset \ldots \ds{IWSLT 2015} \ldots
            \end{varwidth}
        };

            \node[anchor=north west] (caption) at (0.15,7) {Table I: \ldots \ds{test set}\ldots \ms{BLEU} metric.};
            \node [
                below=0.08cm of caption.south west,
                anchor=north west,
            ] (table) {
            \begin{tabular}{cccc}
                \toprule
                & \multicolumn{2}{c}{\ds{WMT 2014}} & \ldots\\
                \ldots & \ds{en-fr} & fr-en & \ldots\\\midrule
                \ldots & \ldots & \ldots & \ldots\\
                NMT (ours) & \textbf{56.3} & 41.8 & \ldots\\\bottomrule
            \end{tabular}
            };
            \node[below=0.2cm of table.south west,anchor=north west, align=left] {
                Linking result:\\
                \small Task: Machine Translation\\
                \small Dataset: WMT2014 English--French Test\\
                \small Metric: BLEU score\\
                \small Value: 56.3\\
                \small Model: NMT\\
                \small Confidence: 0.98
            };
    \end{tikzpicture}

%% file: dataset.tex
\section{Dataset}
\label{sec:dataset}

In this section we explain the datasets we used for training and evaluating \model{} for results extraction. The primary input we use for a training dataset is \LaTeX{} source code of machine learning papers from \arxiv{}. Over 90\% of considered papers have source code available. This allows us to obtain a high quality dataset without common artifacts that arise from extracting data directly from PDF files~\cite{ibm-extraction}.

For training our model we use two main datasets:

\begin{itemize}
  \item \unlabeled{}: An unlabelled dataset of over 100,000 machine learning papers. Used for language model pre-training.
  \item \finegrained{}: A table segmentation dataset where each cell is annotated according to whether it is a paper, metric, dataset, and so on. Used for table segmentation and table type classification.
\end{itemize}

We tune the linking and filtering performance of our model using a validation dataset:

\begin{itemize}
  \item \linkedresults{}: An annotated dataset of over 200 papers with results tuples, capturing the performance of models in the papers, and links to tables.
 \end{itemize}

Lastly we evaluate the end-to-end performance of our model on our test set:

\begin{itemize}
  \item \leaderboards{}: An annotated dataset of over 2,000 leaderboards with results tuples. Used for end-to-end performance evaluation.
\end{itemize}

We now describe in detail these datasets.

\subsection{arXiv Papers}
\label{sec:dataset:unlabeled}

The dataset contains $104,723$ papers published on \arxiv{} between 2007--2020. $94,616$ papers are available with \LaTeX{} sources, from which we extracted $277,996$ tables in total. 
Due to licensing limitations the dataset we release with this paper contains only metadata (available in the public domain) and links to articles. The dataset is unlabeled, designated for use in self-supervised pretraining.

\subsection{Segmented Tables}
\label{sec:dataset:structure}

This is a dataset for table classification and segmentation, containing 1400 annotated tables from 354 articles. The dataset provides data on dataset mentions in captions, the type of table (\texttt{leaderboard}, \texttt{ablation}, \texttt{irrelevant}) and ground truth cell annotations into classes: \texttt{dataset}, \texttt{metric}, \texttt{paper model}, \texttt{cited model}, \texttt{meta} and \texttt{task}.

\subsection{Linked Results}
\label{sec:dataset:results}

This is a set of 236 papers we annotated with 1148 results tuples, capturing the performance of models in the papers. Additionally we include metrics scores in a normalized form. We also record metadata such as the names of the models used in papers. Each results tuple (task, dataset, metric name, metric value) is linked to a particular table, row and cell it originates from. Note that results that appear outside of a table, for instance in the paper's text or graphs, are not present in this dataset.

\subsection{PWC Leaderboards}

This is a dataset of 2,295 leaderboards obtained from the Papers With Code \arxiv{} labelling interface. This interface allows an annotator to take a paper and label it with results tuples. It is therefore a good ground-truth test on which to evaluate the end-to-end performance of our automated solution.
Additionally, each record is linked to a cell it appears in.

%% file: experiments.tex
\section{Experiments}
We now evaluate the end-to-end performance of \model{} on the results extraction task. We evaluate on two datasets: the NLP-TDMS dataset introduced in~\citet{ibm-extraction}, in order to compare our method to the state of the art, and on our \leaderboards{} dataset, which contains many more leaderboards and acts as a more challenging benchmark. 

\subsection{NLP-TDMS Results}

We compare \model{} to the TDMS-IE model from ~\citet{ibm-extraction} on the NLP-TDMS dataset in~\cref{tab:results:end-to-end:nlp-tdms}. The NLP-TDMS (Full) dataset contains 332 papers related to Natural Language Processing with 848 performance annotations of task, dataset, metric and score and 168 unique leaderboards. The subset NLP-TDMS (Exp) is limited to 77 leaderboards appearing in at least 5 papers. See~\cref{tab:datasets:nlp-tdms} in the Appendix for dataset statistics. To compare with~\citet{ibm-extraction}, we use the Exp dataset.

\citet{ibm-extraction} extract records directly from PDF, so the methods are not fully comparable. In order to run~\model{} on that dataset we limit the dataset to papers for which \LaTeX{} source code is available.
\Cref{tab:results:end-to-end:nlp-tdms} shows results on that subset with TDMS-IE performance computed based on published predictions. Our solution yields significantly better results for whole records retrieval despite not being trained on their taxonomy (i.e., the zero-shot scenario in~\citet{ibm-extraction}).

\label{sec:experiments:nlptdms}
\begin{table}[t]
    \centering
    \setlength{\tabcolsep}{4pt}
    \caption{End-to-end extraction results on subset of NLP-TDMS (Exp) dataset.}
    \label{tab:results:end-to-end:nlp-tdms}
    \small
    \begin{tabular}{lcccccc}
     \toprule
  \multirow{2}{*}{Method} & \multicolumn{3}{c}{Micro} & \multicolumn{3}{c}{Macro}\\
  \cmidrule(lr){2-4}\cmidrule(lr){5-7}
   & P & R & F$_1$ & P & R & F$_1$\\\midrule
   \multicolumn{7}{c}{(task, dataset, metric)}\\\midrule
TDMS-IE  & 53.4 & 66.3 & 59.2 & 57.1 & 66.1 & 58.5\\
\model{} & 65.8 & 58.5 & 61.9 & 56.0 & 55.8 & 54.1\\
\midrule
  \multicolumn{7}{c}{(task, dataset, metric, score)}\\\midrule
TDMS-IE  & 6.8 & 8.4 & 7.5 & 8.6 & 9.5 & 8.8\\
\model{} & 27.4 & 24.4 & 25.8 & 20.2 & 20.6 & 19.7\\
     \bottomrule
    \end{tabular}
\end{table}

\subsection{\leaderboards{} Results}
Having validated the performance of our approach compared to the state of the art, we now apply it to our much larger dataset of leaderboards. Compared to the NLP-TDMS dataset, whose taxonomy consists of 77 leaderboards, our taxonomy consists of 3,445 leaderboards making prediction much more challenging. 

The results of our approach for extracting each entity are detailed in~\cref{tab:results:end-to-end:pwc-leaderboards}. We achieve reasonable performance on extracting the full TDMS (task, dataset, metric, score) tuple, which is the most challenging setting and the highest scores for extracting task and metric information. The lower scoring entities are generally the ones that depend on the quality of extraction of other entities. For example, extracting leaderboards depends on how well we extract task, dataset and metric entities.

\begin{table}[t]
    \small
    \centering
    \setlength{\tabcolsep}{4pt}
    \caption{Extraction results of \model{} on  \leaderboards{} dataset (restricted to our taxonomy) for entire records (TDMS), records without score (TDM) and individual entities.}
    \label{tab:results:end-to-end:pwc-leaderboards}
    \begin{tabular}{lcccccc}
     \toprule
  \multirow{2}{*}{Entity} & \multicolumn{3}{c}{Micro} & \multicolumn{3}{c}{Macro}\\
  \cmidrule(lr){2-4}\cmidrule(lr){5-7}
   & P & R & F$_1$ & P & R & F$_1$\\\midrule
TDMS     & 37.4 & 23.2 & 28.7 & 24.0 & 21.8 & 21.1\\\midrule
TDM      & 67.8 & 47.8 & 56.1 & 47.9 & 46.4 & 43.5\\
Task     & 70.6 & 57.3 & 63.3 & 60.7 & 62.6 & 59.7\\
Dataset  & 70.2 & 48.4 & 57.3 & 53.5 & 52.7 & 49.9\\
Metric   & 68.8 & 58.5 & 63.3 & 58.4 & 60.4 & 56.5\\
\bottomrule
    \end{tabular}
\end{table}

%% file: ablation.tex
\section{Performance Studies}
\label{sec:ablation}

In this section, we perform experiments over the various steps of \model{} in order to better understand their relative importance. Our key finding is that table segmentation is the performance bottleneck of \model{}. We run our experiments on the \finegrained{} dataset introduced in~\cref{sec:dataset:structure}.

\subsection{Table Type Classification}
\label{sec:ablation:table-type}

The biggest issue of table type classification is in distinguishing between leaderboard and ablation tables (see~\cref{fig:table-type:cm} in Appendix for the confusion matrix). These tables can be very similar structurally: ablations may even compare on the same split of data as the primary result. As the distinction is not always clear, during results retrieval we extract results from both types of tables and pick only the best result during filtering (i.e., the highest or lowest based on predicted metric).

\subsection{Table Segmentation}
\label{sec:results:structure-prediction}

One goal of table segmentation is to generalise to extract tables from unseen tasks. To study this, we partitioned \finegrained{} dataset into 11 folds, based on the task name extracted from paper abstracts. The fold with tables from Image Classification papers is always used as a validation set. For each of the remaining 10 folds we train 5 models with a given fold used as a test set and the other 9 folds used as training data. The final table segmentation model used in \model{} is the one with the highest micro F$_1$ score on the validation set.

\Cref{tab:results:segmentation:kfolds} shows micro precision and recall of classifying each non-numeric cell into one of 5 exclusive classes: dataset, metric, competing model, paper’s model or other.

We can see that we achieve strong results on all tasks, although some tasks perform better than others. A task like semantic segmentation has less table and benchmark diversity, so benchmark tables for datasets like Cityscapes and PASCAL VOC 2012 are fairly standardised across papers. This makes extraction fairly straightforward. In contrast, the worse performing tasks are unusual in their own way. In image generation, for instance, we are less able to extract the correct dataset entity, whereas in speech recognition, our model has more problems distinguishing paper models from competing models; see~\cref{fig:segmentation:cm} in the Appendix.

\begin{table}[t]
\small
    \centering
    \setlength{\tabcolsep}{4pt}
    \caption{Table segmentation results for 10-fold training with image classification papers fixed as a validation set and variable test set. Micro precision, recall and F$_1$ score are averaged over 5 runs.}
    \label{tab:results:segmentation:kfolds}
    \begin{tabular}{lrrrrrr}
     \toprule
     & \multicolumn{3}{c}{validation} & \multicolumn{3}{c}{test}\\
     \cmidrule(lr){2-4}\cmidrule(lr){5-7}
 test set & P & R & F$_1$ & P & R & F$_1$ \\\midrule
image gen.       &       84.5 &    87.9 & 86.2 &       73.4 &    81.6 & 77.3 \\
misc.            &       84.0 &    88.2 & 86.0 &       81.7 &    93.5 & 87.2 \\
machine trans.   &       83.1 &    90.8 & 86.8 &       80.5 &    94.4 & 86.9 \\
NLI              &       83.6 &    89.6 & 86.5 &       84.5 &    97.3 & 90.4 \\
object detection &       81.9 &    91.4 & 86.3 &       83.7 &    96.7 & 89.7 \\
pose estimation  &       85.1 &    89.9 & 87.4 &       86.0 &    96.8 & 91.1 \\
question ans.    &       83.6 &    89.5 & 86.4 &       80.4 &    89.6 & 84.8 \\
semantic seg.    &       81.4 &    91.1 & 86.0 &       90.2 &    95.9 & 92.9 \\
speech rec.      &       84.7 &    89.8 & 87.2 &       67.2 &    90.7 & 77.1 \\
text class.      &       83.9 &    90.4 & 87.0 &       74.9 &    93.3 & 83.1 \\
\bottomrule
    \end{tabular}
\end{table}

\subsection{Linking}
\label{sec:ablation:linking}
To evaluate linking performance in isolation of other steps we run it on tables with ground truth type and segmentation annotations. The annotations are available in the~\finegrained{} dataset for 24 Speech Recognition and 32 Semantic Segmentation papers with 287 annotated leaderboard records in total. For each cell with associated leaderboard annotation we generate cell contexts and use linking to retrieve the top-5 predictions.
We test four approaches to generate evidence of mentions.
\paragraph{Bag-of-Words} The full name and any word (which is not an English stop-word) occurring in the name of a metric or dataset (as found in taxonomy) is evidence of mention.
\paragraph{Abbreviations} We run an abbreviation detector~\cite{scispacy} over the~\unlabeled{} dataset to extract pairs of common abbreviations and their full forms. The previous approach is extended with abbreviations of full forms occurring in name of metric or dataset. For example, with a pair (\textit{en-vi}, \textit{English-Vietnamese}) and dataset name \textit{IWSLT2015 English-Vietnamese}, \textit{en-vi} is added as mention evidence for this dataset.
\paragraph{Manually Curated} We extend the Bag-of-Words approach with list of manually curated mention evidence. Only mentions of datasets and metrics related to speech recognition and semantic segmentation are modified.
\paragraph{Combined} The previous approach extended with abbreviations.\\

In~\cref{tab:results:linking:gold} we show Top-1 and Top-5 accuracy of the predictions over all leaderboard records from each collection of papers. Using abbreviations significantly improves the performance over bag-of-words approach. The worse performance caused by adding abbreviations to manually curated lists suggests that abbreviations could increase rate of false-positive matches of mentions. Another explanation is that manually curated lists of mentions could be biased towards leaderbords related to speech recognition and semantic segmentation due to construction of the lists.

The overall performance of the linking step allows us to use it in production environment for efficient semi-automated extraction of results. Our solution proposes to users the Top-5 predictions associated with cells they pointed, thus eliminating the tedious and error-prone step of matching the results with existing leaderboards and ensuring that metric values are correctly normalized.

\begin{table}[t]
    \centering
    \setlength{\tabcolsep}{3.5pt}
    \caption{Linking performance using ground truth annotations of table types and segmentation. }
    \small
    \label{tab:results:linking:gold}
\begin{tabular}{lcccccccc}
\toprule
& \multicolumn{8}{c}{Top-1 Accuracy [\%]}\\
\midrule
\multirow{2}{*}{evidence} & \multicolumn{4}{c}{speech rec.} & \multicolumn{4}{c}{sem.\ segmentation} \\
&                TDMS & T & D & M &                   TDMS & T & D & M \\
\cmidrule(lr){2-5}\cmidrule(lr){6-9}
BoW &                 42 &   86 &      45 &     72 &                    49 &   95 &      71 &     67 \\
abbrs    &                 56 &   87 &      57 &     74 &                    56 &   95 &      79 &     74 \\
curated  &                 76 &   87 &      77 &     87 &                    77 &   95 &      89 &     87 \\
combined &                 67 &   87 &      68 &     78 &                    72 &   95 &      86 &     85 \\
\midrule
& \multicolumn{8}{c}{Top-5 Accuracy [\%]}\\
\midrule
\multirow{2}{*}{evidence} & \multicolumn{4}{c}{speech rec.} & \multicolumn{4}{c}{sem.\ segmentation} \\
&                TDMS & T & D & M &                   TDMS & T & D & M \\
\cmidrule(lr){2-5}\cmidrule(lr){6-9}
BoW &                 72 &   88 &      73 &     84 &                    82 &   99 &      89 &     93 \\
abbrs    &                 76 &   89 &      76 &     84 &                    93 &  100 &      94 &     99 \\
curated  &                 85 &   90 &      85 &     91 &                    97 &   99 &      99 &     99 \\
combined &                 81 &   89 &      81 &     89 &                    97 &   99 &      99 &     99 \\
\bottomrule
\end{tabular}
\end{table}

%% file: conclusions.tex
\section{Future Work}

We cover three possible extensions to our work for future research. 

First, we might want to consider methods that retrieve \textit{\textbf{all}} results rather than just the principal results introduced in the paper. This includes extracting ablation studies to enable search over fine-grained comparison results. 

Secondly, we could look more into automatic taxonomy discovery. Currently, we assume a closed-domain approach with taxonomy of leaderboards known in advance. While manually extending the taxonomy requires only adding the task, dataset and metric names, it becomes problematic to cover large fraction of the papers due to publication rate and long tail of leaderboards. 

Finally, to relax the necessity of~\model{} to have access to \LaTeX{} source we consider using the~\unlabeled{} dataset as a corpus to train extraction working directly with PDF files.

\section{Conclusions}
We presented an end-to-end model for extracting results from machine learning papers. Our method performs well across various tasks and leaderboards within machine learning, with a taxonomy that can be easily extended without retraining. Additionally we released a new collection of datasets for training and evaluating on the results extraction task. These datasets enable the training of more fine-grained feature extractors and detailed error analysis. We demonstrated that our approach achieves significant performance gains over the state-of-the-art. Future work may want to build on our approach for more comprehensive extraction tasks, focussing on more types of result, as well as other information contained in papers such as architectural details and hyperparameters.

%% file: acknowledgements.tex
\section*{Acknowledgements}

The authors would like to thank Waleed Ammar, Sebastian Kohlmeier and Iz Beltagy on useful discussion and feedback. 

%% file: appendix.tex
\newpage
\clearpage
\appendix
{\noindent\Large\textbf{Appendix}}
\section{Training Details}
\begin{table}[H]
    \centering
    \setlength{\tabcolsep}{8pt}
    \caption{ULMFiT architecture and hyperparameters used for table type classification and table segmentation.}
    \label{tab:ulmfit}
    \begin{tabular}{lr}
     \toprule
     vocabulary size & 30K\\
     tokenization & unigram model\\
     RNN type & LSTM\\
     recurrent layers & 3\\
     embeddings dimension & 400\\
     hidden state dimension & 1152\\
     pretraining & 12 epochs\\
     batch size & 256\\
     \bottomrule
    \end{tabular}
\end{table}

\begin{table}[H]
\tiny
    \centering
    \setlength{\tabcolsep}{4pt}
    \caption{Features For Table Segmentation}
    \label{tab:table-structure:features}
    \begin{tabular}{l|p{5cm}}
     \toprule
     Feature & Description\\\midrule
is emphasised & whether text in cell is boldfaced, colored, etc.\\
cell style & e.g. "align-left top-border"\\
text & mentions of cell’s content (as in \cref{fig:contexts})\\
cell content & cell's content without styles and references, e.g. “ULMFiT” \\
row context & concatenated cell's row, e.g. "ULMFiT <sep> 94.5\% <sep> 92.1\% <sep>" \\
column context & concatenated cell’s column, e.g. “Method <sep> LSTM <sep> GRU <sep> ULMFiT <sep> BERT”\\
cell reference & list of reference ids used in cell, e.g. “bib4, bib18”\\
\bottomrule
\end{tabular}
\end{table}

\section{Datasets}
\begin{table}[H]
    \centering
    \setlength{\tabcolsep}{4pt}
    \caption{Statistics of the NLP-TDMS Full and Exp datasets.}
    \label{tab:datasets:nlp-tdms}
    \begin{tabular}{lrr}
     \toprule
 & Full & Exp\\\midrule
     unique taxonomy entries & 168 & 77\\
     unique tasks & 35 & 18\\
     unique datasets & 99 & 44\\
     unique metrics & 72 & 30\\\midrule
 papers & 332 & 332\\
     results & 848 & 606\\
     \bottomrule
    \end{tabular}
\end{table}

\begin{table}[H]
    \centering
    \setlength{\tabcolsep}{8pt}
    \caption{Statistics for the \leaderboards{} dataset with all entries (Full) and entries restricted to our taxonomy (Restricted).}
    \label{tab:dataset:labeled}
    \begin{tabular}{lrr}
     \toprule
     & Full & Restricted\\\midrule
    unique taxonomy entries   &  2295 &   649\\
    unique tasks              &   252 &   134\\
    unique datasets           &  1156 &   433\\
    unique metrics            &   414 &   162\\\midrule
    papers                    &   733 &   516\\
    results                   &  5406 &  2802\\
     \bottomrule
    \end{tabular}
\end{table}

\section{Additional Results}

\begin{figure}[H]
\centering
\caption{Confusion matrices of segmenting cells into five classes: dataset (including subdatasets), metric, model introduced in processed paper, competing model and other. Results averaged over 5 runs for each task, using 10-fold training as described in~\cref{sec:results:structure-prediction}.}
\label{fig:segmentation:cm}
\begin{subfigure}{0.38\textwidth}
\centering
\includegraphics[width=\linewidth]{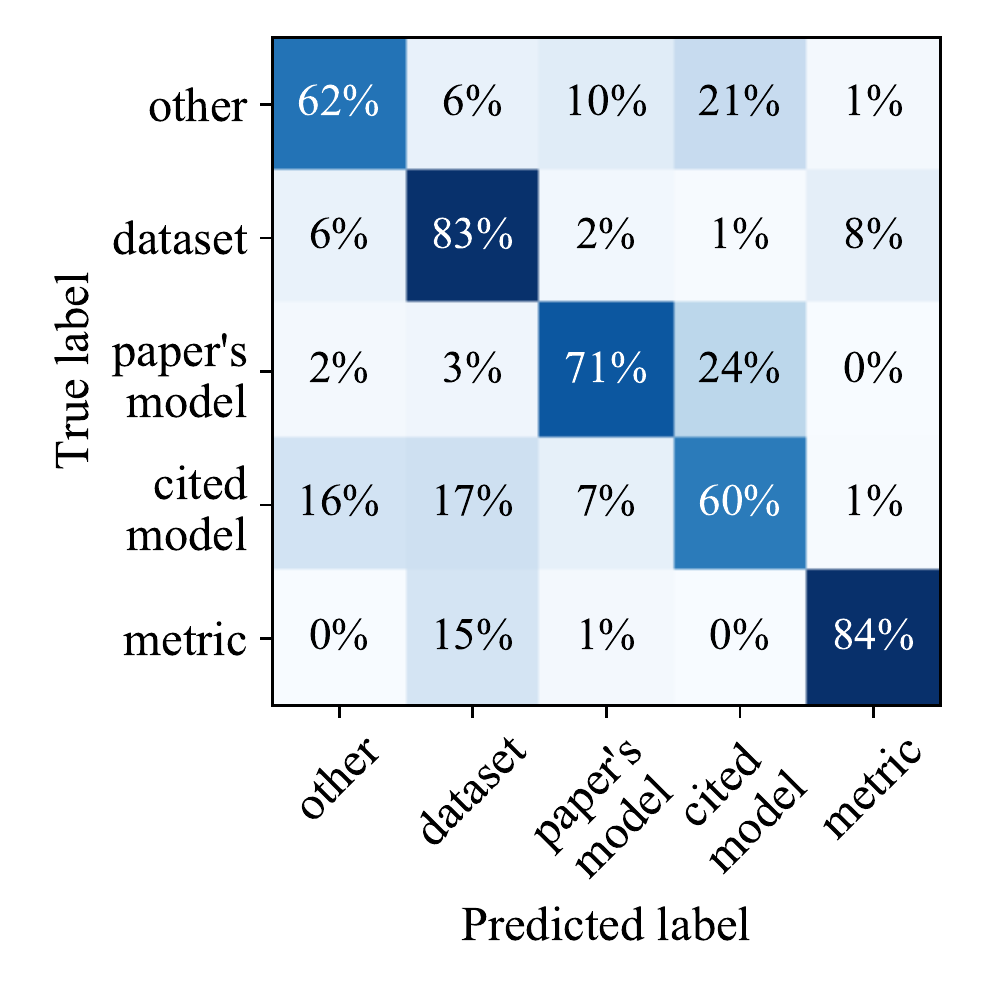}
\caption{Speech Recognition}
\end{subfigure}
\begin{subfigure}{0.38\textwidth}
\centering
\includegraphics[width=\linewidth]{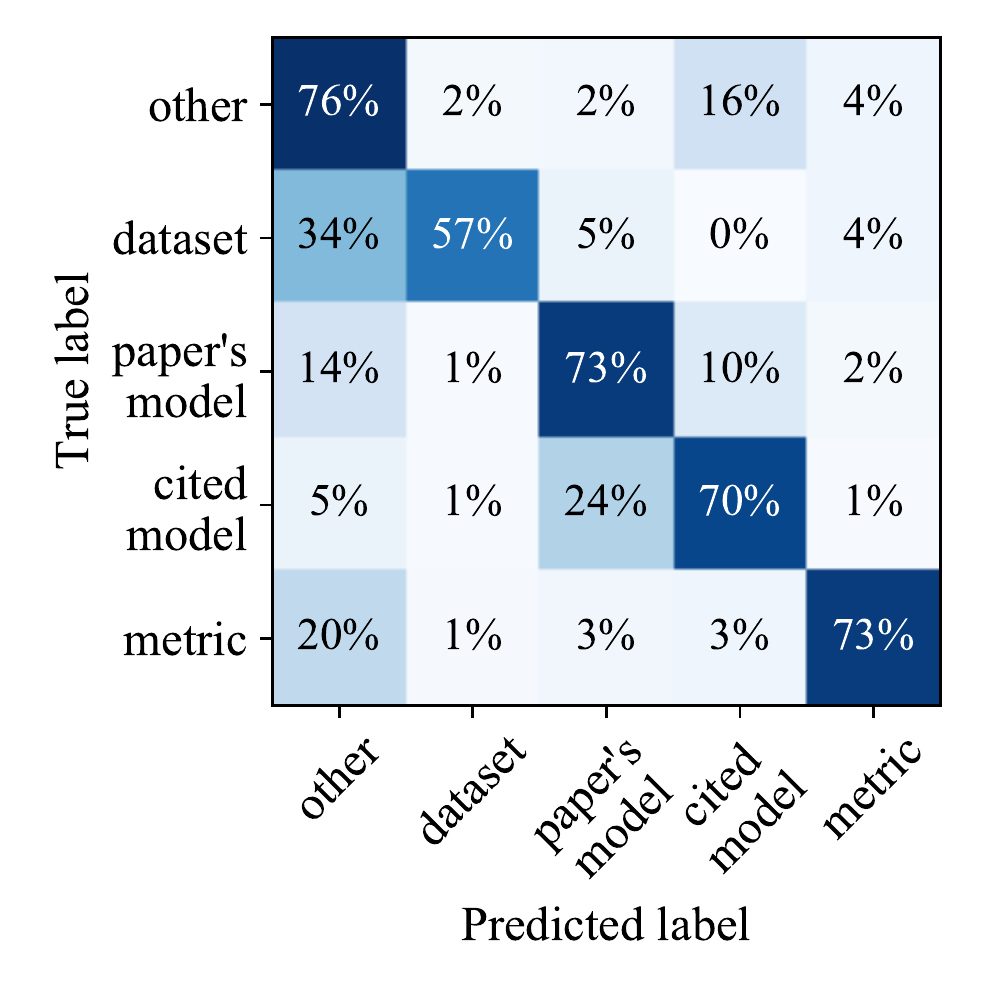}
\caption{Image Generation}
\end{subfigure}
\begin{subfigure}{0.38\textwidth}
\centering
\includegraphics[width=\linewidth]{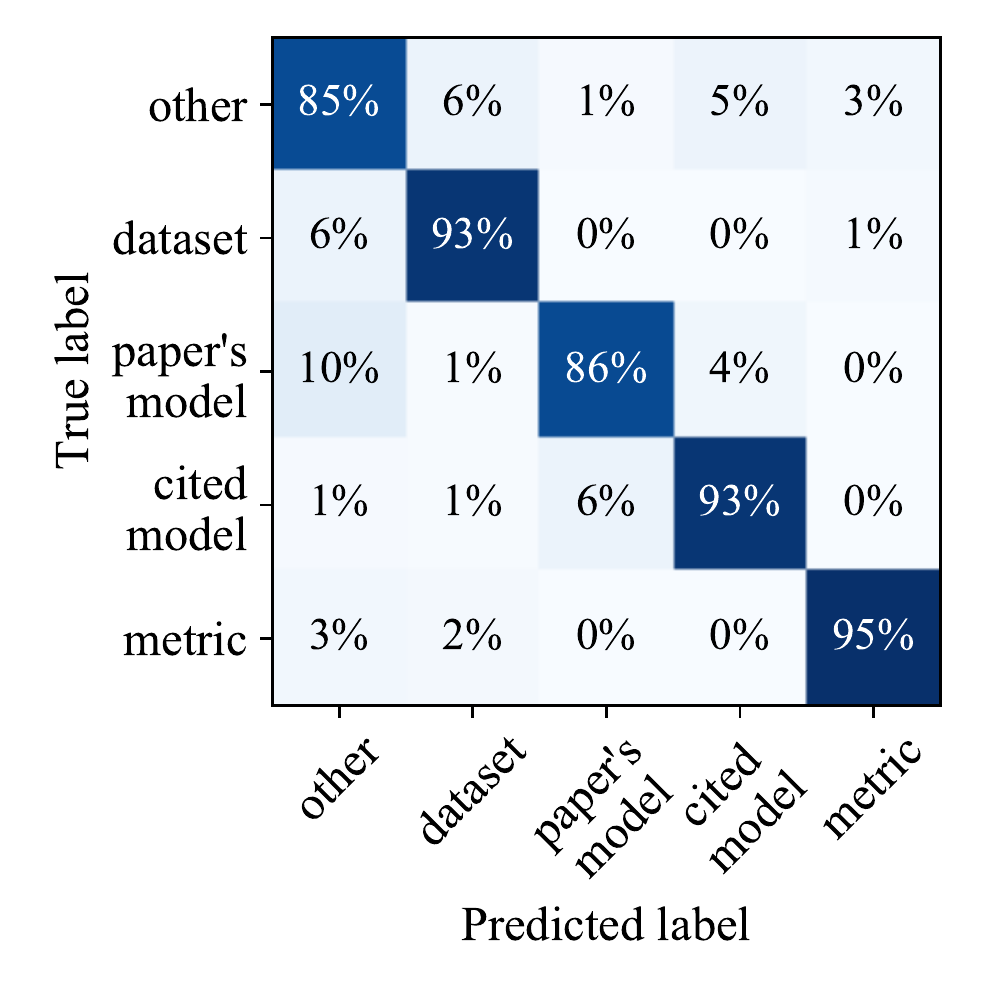}
\caption{Semantic Segmentation}
\end{subfigure}
\end{figure}

\begin{figure}[t]
\centering
\caption{Confusion matrix of table type classification step.}
\label{fig:table-type:cm}
\includegraphics[scale=0.9]{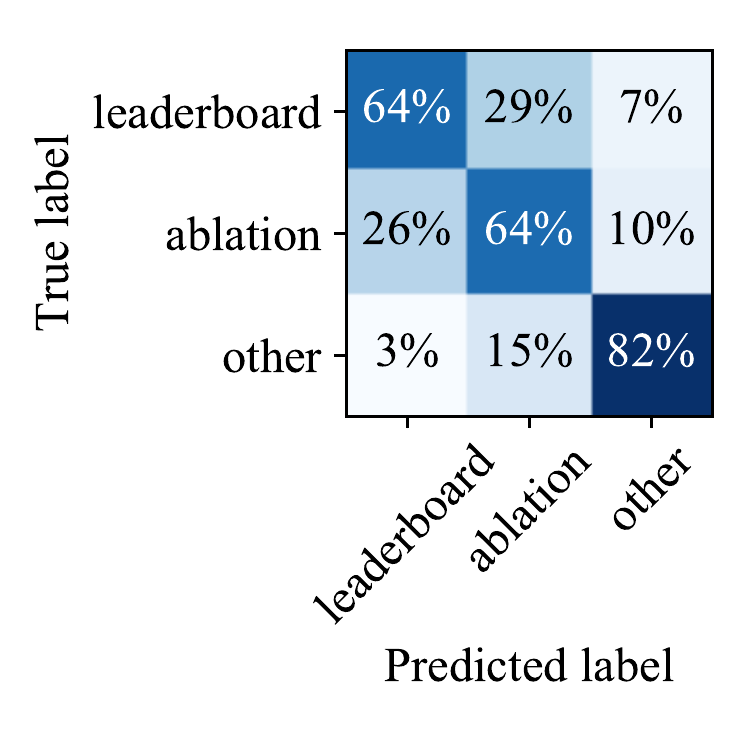}
\end{figure}

\section{Mention Probabilities}
\label{sec:mention-probs}

Using the methodology from~\cref{sec:linking-inference}, we can calculate $\prCond{y_{k}}{E}$ by combining mentions $\prCond{\textit{mention}}{y_{k}^{\textit{(entity)}}}$. We simplify the notation by referring to this conditional distribution as $\prCond{m}{f}$ in this section. This denotes the probability that a mention evidence $m$ for given \textit{entity} $f$ appears in a particular context of cell containing results for leaderboard with entity $f$.

We compute all possible mentions directly from tasks, datasets and metrics names appearing in leaderboards. For a name of dataset or metric the mentions list consists of the whole name as well as each word, without duplicates and English stop words. As tasks names often consist of common words, to limit the number of false positives the mentions list for a given task contains only that task's name.

We compute probability $\prCond{m}{f}$ assuming all mentions (separately for tasks, datasets and metrics) are distributed uniformly, $\prCond{f_1}{m} = \prCond{f_2}{m}$ for all $f_1, f_2$ for which $m$ is a mention evidence. We then use Bayes rule to get $\prCond{m}{f}$, assuming that all mentions of a given type are distributed uniformly. This results in the conditional probability of a mention being inversely proportional to the number of entities having that mention evidence in common:
\begin{equation*}
    \prCond{m}{f} \propto \frac{1}{| \{ g: m \textit{ is mention evidence for }g\} |}.
\end{equation*}